\documentclass[journal]{IEEEtranTIE}
\usepackage{graphicx}
\usepackage{cite}
\usepackage{picinpar}
\usepackage{amsmath}
\usepackage{amssymb}
\usepackage{amsthm}
\usepackage{multirow, booktabs}
\usepackage{txfonts}
\usepackage{url}
\usepackage{flushend}
\usepackage[latin1]{inputenc}
\usepackage{colortbl}
\usepackage{soul}
\usepackage{multirow}
\usepackage{pifont}
\usepackage{color}
\usepackage{alltt}
\usepackage[hidelinks]{hyperref}
\usepackage{enumerate}
\usepackage{siunitx}
\usepackage{breakurl}
\usepackage{epstopdf}
\usepackage{pbox}
\usepackage[T1]{fontenc}
\usepackage{threeparttable}

\usepackage{etoolbox}
\makeatletter
\patchcmd{\@makecaption}
{\scshape}
{}
{}
{}
\makeatother
\makeatletter

\newcommand{\Rmnum}[1]{\expandafter\@slowromancap\romannumeral #1@}
\makeatother
\begin{document}
	\title{Domain Knowledge integrated for Blast Furnace Classifier Design }
	
	\author{
		\vskip 1em
		
		Shaohan Chen,
		Di Fan, and
		\\  Chuanhou Gao, \emph{Senior Member, IEEE}
		
			
			
			
	}
	
	\maketitle
	
	\begin{abstract}
		Blast furnace modeling and control is one of the important problems in the industrial field, and the black-box model is an effective mean to describe the complex blast furnace system. In practice, there are often different learning targets, such as safety and energy saving in industrial applications, depending on the application. For this reason, this paper proposes a framework to design a domain knowledge integrated classification model that yields a classifier for industrial application. Our knowledge incorporated learning scheme allows the users to create a classifier that identifies "important samples" (whose misclassifications can lead to severe consequences) more correctly, while keeping the proper precision of classifying the remaining samples. The effectiveness of the proposed method has been verified by two real blast furnace datasets, which guides the operators to utilize their prior experience for controlling the blast furnace systems better.
	\end{abstract}
	
	
	\markboth{IEEE TRANSACTIONS ON INDUSTRIAL ELECTRONICS}%
	{}
	
	\definecolor{limegreen}{rgb}{0.2, 0.8, 0.2}
	\definecolor{forestgreen}{rgb}{0.13, 0.55, 0.13}
	\definecolor{greenhtml}{rgb}{0.0, 0.5, 0.0}
	
	\section{Introduction}
	\IEEEPARstart{A}{blast furnace} is a metallurgical reactor used to smelt and produce molten iron, commonly known as molten iron or pig iron. It is a tall chimney structure. A certain amount of metal ore, coke and flux are loaded on the top of the chimney, and then hot air is blown into the bottom to compress the air. When the solid material moves downward and the hot combustion gas flows upward, the whole furnaces take place both chemical reactions and heat transfer phenomena. A large amount of heat energy is generated during the ironmaking process, which can raise the temperature of the blast furnace to nearly $2000^\circ \mathrm{C}$. The final product, consisting of slag and hot metal, sinks to the bottom and is periodically released for subsequent refining. One ironmaking cycle takes about 6-8 hours.
	
	For many countries, the metallurgical industry plays an important role in the national economy, so there is widespread interest in blast furnace modeling and control. 
	Zhou et al.\cite{zhou2020data} propose a novel integrated PCA-ICA method for monitoring and diagnosing the abnormal furnace conditions in blast furnace ironmaking. Since the hot metal silicon content simultaneously reflects the product quality and the thermal state of the blast furnace, in order to facilitate the realization of control, Li et al.\cite{li2017bayesian} proposes a Bayesian block structure sparse based Takagi-Sugeno (T-S) fuzzy modeling method. Also, a two-stage online prediction method based on an improved echo state network is proposed to realize forecasting in the blast furnace gas system\cite{zhao2010two}. Huang et al.\cite{huang2020two} presents a two-stage decision-making method
    for the burden distribution parameters based on recognizing the conditions. Thanks to the rapid development of machine learning, the relevant technology techniques have rapidly helped us make decisions or find the optimal control policies for complex the iron-making processes \cite{gao2014rule}. However, lack of theoretical interpretability for many machine learning methods has been an obstacle to applying them with confidence. Also, it is difficult to adjust or design these models with deep structures to achieve some particular learning targets (i.e. safety and energy saving) in industrial applications. Thus, it is necessary to build a \textbf{data-efficient} learning scheme to adapt for the special requirements of industrial applications.
	
	To enhance model's interpretability, and achieve particular learning targets,  incorporating the domain experts' experience into the black-box ML models (with the shallow structures) is a feasible strategy to make the produced models perform better with the same size of the data. The benefits of incorporating prior information into the ML models has been verified in \cite{niyogi1998incorporating}, \cite{lauer2008incorporating}. In particular, Mangasarian et al. \cite{mangasarian2004knowledge}, \cite{mangasarian2007nonlinear},
	\cite{mangasarian2008nonlinear} considered to impose the information of boundary of the classification hyperplane on the SVMs to get higher accuracy. However, it is still unclear which types of the prior knowledge can improve the model performance and how to incorporate prior knowledge into the black-box models suitably.
	This paper aims to explore the effectiveness of incorporating domain knowledge into the black-box models in the context of solving the control problem of blast furnace. To be specific, we wish to design a knowledge integrated classification model such that the yielded classifier to meet the "safety" and "energy saving" demand of the industrial applications. 
  
	To naturally define our learning target, we consider the silicon prediction tasks in real blast furnace systems. The silicon content is thought as an essential index to reflect the inside temperature distribution of the blast furnace. Abnormally low silicon content often indicates the significant drop of the temperature of the blast furnace, which results in severe damages to the blast furnace equipments. So, failing to identify the low silicon samples (especially when the last record already showed the abnormal silicon evolution) will be more dangerous. Also, an unusually high silicon content often indicates a significant increase in blast furnace temperatures, which requires excessive energy consumption. Therefore, failure to identify high silicon samples means higher energy consumption\cite{jian2012binary}. Therefore, for machine learning systems to achieve learning targets in this real industrial applications, it is critical to build a classifier that can optimally trade off different kinds of risks. In other words, we want to create a learning target which has \textbf{risk preferences} according to the \textbf{domain experts' knowledge} in the industrial application areas. Given the analysis above, we measure the "performance" of a classification model as the weighted sum of accuracies for different classes of data, which is induced from the practical industry goals. Or more formally, we refer to the expectation of ensemble negative risks, which will be treated as the learning target in this context. More details can be seen in section \Rmnum{4}.

	There have been many classifiers design methods for achieving specific learning purposes. Classical methods, such as standard support vector machines (SVMs) in both offline \cite{cortes1995support} and online \cite{kivinen2004online}, \cite{cauwenberghs2001incremental} settings, are designed for classifying problems with an equal cost for both classes and they have been very effective on several application problems. 
	However, the use of these models is limited, for example, when the data is highly unbalanced. Lin et al. \cite{lin2002support} modified the standard SVMs adapting for the unbalanced classification task.
	Heuristic classifier design methods, for example, 
	Gao et al. \cite{gao2013one} introduced a one-pass optimal AUC (Area under the ROC curve) method where the data is highly unbalanced.
	Xu et al. \cite{xu2014classifier} presented an algorithm to  balance the performance with test-time cost efficiently. 
	Lin et al. \cite{lin2011design} designed a privacy-preserving SVM classifier to protect the privacy of data. 
	Wu et al. \cite{wu2004incorporating} proposed a weighted SVM model to combine with some prior knowledge represented by the weights of every training data point, etc. In this paper, we create a new knowledge integrated model to satisfy the requirement for the industrial applications. And our experimental results indicate that incorporating domain knowledge into the SVMs can achieve better performance than the model presented by Lin et al. \cite{lin2002support} given the same size of the training data. Thus, our knowledge integrated method is data-efficient. 
 
	The contributions of this paper can be concluded as follows. \begin{enumerate}[1)]
		\item The proposed knowledge integrated method provides a general framework for embedding domain knowledge into the black-box models to achieve the safety and energy saving requirement for industrial applications. And the model doesn't need to sample extra data points outside the given training dataset. 
		\item To the best knowledge of the authors, this is the first work to verify the possibility of utilizing domain knowledge to improve the black-box model performance on two real blast furnace datasets when the number of training data is fixed.
	\end{enumerate}

 	The rest of the paper is organized as follows.  
	Section \Rmnum{2} presents the blast furnace problem. Second \Rmnum{3} introduce the basic properties of the standard SVM and the variant of SVM which uses the unequal misclassification costs. Then, the whole framework to design the knowledge integrated classification model to achieve the specific learning target are introduced in Second \Rmnum{4}. We perform case study based on two real blast furnace datasets to illustrate the effectiveness of the proposed knowledge incorporated model in section \Rmnum{5}. Finally, section \Rmnum{6} concludes this paper and presents the future work.

	\section{BLAST FURNACE PROBLEM}
	One of the main tasks of blast furnace control is to control the hot metal temperature and composition, like the silicon content in the hot metal within acceptable bounds. The hot metal silicon content indicates the in-furnace thermal status since the silicon transfer from silica to the hot metal occurs as an endothermic reaction that could affect the bottom of the furnace (i.e., the hearth) and further influence the hot metal temperature \cite{chen2019linear}. On the other hand, the change of silicon content reflects the consumption of coke in the raw material, that is, the increase of silicon content often means the excess of coke, while the decrease of silicon content means the consumption of coke. For safety and benefits in the iron-steel company, it is desired to operate the blast furnace at the proper hot metal silicon content, and more importantly, avoid the risk of chilled hearth \cite{saxen2012data} and high energy consumption, i.e., the hot metal silicon content is not too low and not too high.  
	
	Let $z$ represent the hot metal silicon content in this context. To prevent the happening of the chilled hearth, the acceptable infimum of $z$ is denoted as $z_{inf}$. And, in order to prevent the high energy consumption and the serious impact of high temperature in the hearth, the acceptable supremum value of z is represented by $z_{sup}$. We aim at manipulating the blast furnace inputs so that the hot metal silicon content falls into the range $[z_{inf},z_{sup}]$. Therefore, we can transform the blast furnace control problem into a three-class classification task, then construct a classifier to separate these three types of samples correctly, and finally provide the suggestions for the blast furnace control. Specifically, we label the data whose outputs lie in the interval $[0, z_{inf})$ as the "low silicon", in $[z_{inf},z_{sup}]$ as the "proper silicon", and in $(z_{sup}, 1]$ as the "high silicon". In the following, we will show the capacity of the proposed knowledge integrated classifier to separate these three types of data using two real blast furnace datasets.
	
	\section{PRELIMINARY ON SOFT-MARGIN SVM AND ITS EXTENSIONS}
	In this section, we introduce the basic concepts of the soft-margin SVM  and then illustrate several typical examples which serve as an excellent source of inspiration for building our learning scheme. 
SVM is a  kernel-based black-box modeling method, the main idea of which is to construct a hyperplane in an imaginary high-dimensional feature space that could separate two different classes (labeled by the output $y=+1$ or $y=-1$) as far as possible \cite{cortes1995support}.The main SVM model does not allow misclassification of training samples, such SVM is identified by hard-margin. To handle the case of indivisible training data, Sch\"olkopf and Smola \cite{scholkopf2002learning} propose a soft-margin version, which introduces non-negative slack variables to measure the degree of misclassification. Now suppose we are given a training dataset of $m$ examples  $\left( \textbf{x} _ { i } ,y _ { i } \right) _ { i = 1} ^ {m}$, the original soft-magin SVM is the following optimization problem: 
 	\begin{eqnarray}\label{original SVM}
		\begin{array}{cl}
			\min_{w ,b ,\xi_{i}\in \mathbb{R}} &  \frac { 1} { 2} w^{T}w+\frac { C } { m } \sum _ { i = 1} ^ { m } \xi _ { i }\\
			
			\text{s.t.} & y_i(w^T\Phi(\textbf{x}_i)+b)\geq
		1- \xi _ { i } \\
			& \xi _ { i } \geq 0 ,\text{ for } i = 1,\dots ,m
		\end{array} 
	\end{eqnarray}
	
 Here, $w$ is the normal vector in the high-dimensional feature space, $b \in R$ the offset, $\xi_i> 0$ the nonnegative slack variable,  $T$ denotes transpose, $\Phi$ the high-dimensional feature , $C > 0$ is a trade-off parameter which may depend on $m$ .  From \cite{bishop2006pattern}, we know that for soft margin SVM, the slack variable $\xi _ { i }\geq 0$ where i=1,...,m, with one slack variable for each training data point. These are defined by $\xi _ { i }= 0$ for data points that are on or inside the correct margin boundary and $\xi _ { i }=  \left| y_i-f(x_i)\right|={(1-y_if(x_i))}_+$  for other points. Thus a data point that is on the decision boundary $f(x_i)=0$ will have $\xi _ { i }=1$, and points with $\xi _ { i }\geq 1$ will be misclassified. 

	In this original SVM model, we assumed that the cost of different types of misclassification is the same. However, many real-world situations are non-standard, and the most common non-standard hypothetical behavior is that different types of misclassification may have different costs, i.e. one type of misclassification tends to be more severe than the other\cite{lin2002support}, so we should take this into account when building our classification model. Lin et al. \cite{lin2002support} considered the non-standard situation by integrating the unequal misclassification costs (without consideration of sampling bias here) with SVM. They built a  model (\ref{svm2}) by setting the costs for false negative and false positive are $c ^ { + }$ and $c ^ { - }$ ($c^{-}\not=c^{+}$), respectively. 
  	\begin{eqnarray}\label{svm2}
		\begin{array}{cl}
			\min_{w ,b ,\xi_{i}\in \mathbb{R}} &  \frac { 1} { 2} w^{T}w+\frac { C} { m } \sum _ { i = 1} ^ { m }L(y_{i}) \xi _ { i }\\
			
			\text{s.t.} & y_i(w^T\Phi(\textbf{x}_i)+b)\geq
		1- \xi _ { i } \\
			& \xi _ { i } \geq 0 ,\text{ for } i = 1,\dots ,m
		\end{array} 
	\end{eqnarray}
	Here, $L(\cdot)$ is a two-valued function where $L(-1)=c^+$ and $L(+1)=c^-$, respectively. By introducing Lagrangian multipliers $\alpha=(\alpha_1, \alpha_2, \alpha_N)\in R^N_+$, kernel trick, and dual transformation, the optimization problem may be rewritten as follows:
 \begin{eqnarray}\label{modified svm3}
		\begin{array}{cl}
			\min_{\alpha} & \frac{1}{2}\sum_{i=1}^{m}\sum_{j=1}^{m}\alpha_i\alpha_jy_iy_jK(\textbf{x}_i,\textbf{x}_j)-\sum_{i=1}^{m}\alpha_i\\
			
			\text{s.t.} & \sum_{i=1}^{m}\alpha_iy_i=0 \\
			& 0\leq \alpha_i \leq {\frac{C}{m}L(y_i)},\text{ for } i = 1,\dots ,m
		\end{array} 
	\end{eqnarray}
    Here, $K(\textbf{x}_i, \textbf{x}_j)$ is the  kernel function \cite{cristianini2000introduction} defined by 
    \begin{eqnarray}\label{kernel}
		 K(\textbf{x}_i,\textbf{x}_j)=\Phi(\textbf{x}_i)^T\Phi(\textbf{x}_j)
	\end{eqnarray}
    The decision function for model (\ref{svm2}) follows:
    \begin{eqnarray}\label{decision function}
		 f(\textbf{x})=sign(\sum_{\alpha_i\geq 0}^{i=1\sim N}\alpha_i y_i K(\textbf{x}_i,\textbf{x}_j)+b)
	\end{eqnarray}
    Notice that the original SVM represented by Eq. (\ref{original SVM}) is a model in which costs for different classes are equal. However, the model Eq. (\ref{svm2}) implies costs for different classes are $c^+$ and $c^-$ respectively. Moreover, if $c^+$ is larger than $c^-$, then the cost of the classifier misclassifying the negative class is greater than the cost of misclassifying the positive class, and vice versa.
	
	In this paper, we aim to build a general learning framework for designing a knowledge incorporated model that adapts to the specific learning purpose, i.e., satisfying the requirement for industrial applications. Therefore, inspired by the above models, we next introduce the prior knowledge we utilize, and then build a classification model that combine the prior knowledge with the classifier in a way that assigns different misclassification error rates to different samples, ultimately achieving practical industrial goals.

\section{CLASSIFIER DESIGN BASED ON THE DOMAIN KNOWLEDGE}
	In this section, the domain knowledge will be introduced and integrated into soft-margin SVM model to achieve our particular learning target in the special industrial background.
\subsection{Domain knowledge}
	Now we consider the silicon prediction tasks in real blast furnace systems. The silicon content is thought as an important index to reflect the in-furnace thermal status of the blast furnace. Abnormally low silicon content often means the significant drop of the temperature of the hot metal which results in serious damages to the blast furnace equipments, i.e., the chilled hearth. So, failing to identify the "low silicon" sample, especially for the one whose silicon content of the last record was already low (defined as the region $\mathcal{A}$), is even more dangerous. Similarly, failing to identify the "high silicon" sample, especially for the one whose silicon content of the last record was already high, should also be paid attention to. Therefore, we desire to build a classification model that efficiently identifies important samples which will cause severe consequences when missing recognize them, while keeping the proper precision with respect to the remaining samples. For model (\ref{svm2}), it has different penalties for different types of data points, but here we need to consider the silicon content information at the previous moment of the blast furnace hot metal in addition to the information of different classes of data points here. Therefore, based on the idea of model (\ref{svm2}), we propose to add prior information in the similar way into the classification model to achieve this goal.

     In machine learning fields, prior knowledge can refer to any known information about or related to the concerning objects, such as data, knowledge, specifications, etc., \cite{qu2011generalized}. There are mainly two types pf prior knowledge in classification problems: class-invariance and knowledge on the data \cite{lauer2008incorporating}. In this work, we mainly focus on the second type of prior information extracted by the domain experts' experiences which differs from the knowledge mined from the data studied by \cite{chen2019linear}, and with the expression 
	\begin{equation}\label{prior1}
		g({x})\leq 0 \Longrightarrow  \text{The cost to misclassify sample }x ~\text{is} ~\hat{c},~~\forall~ {x}\in \Gamma^+
	\end{equation}
	where $\Gamma^+ :=\{x: \text{label}(x)=+1\}$, $g:\Gamma^+ \to \mathbb{R}$, and $\hat{c}>1$. Here, we consider binary classification problem, labeled by "+1" and "-1" respectively, and assume that samples with positive labels are more "important" than the negative ones, which means that the positive class samples are more urgent than the negative ones to be classified right. Intuitively, we can understand this piece of logical knowledge as misclassifying any positive class sample in the region of $\mathcal{A}=\left\{x|g({x})\leq{0}\right\}$ will suffer a big risk weighted by value $\hat{c}$. Here, $\hat{c}$ can take different values according to different learning targets in real applications.
	
Once we have the form of the prior information, we can assume that for each set of blast furnace data, the prior knowledge of the two classifiers corresponds to the following two expressions, respectively.
	\begin{equation}\label{prior2}
		\begin{split}
			q^{-1}z\leq z_{inf} \Longrightarrow &  \text{The cost to misclassify sample }x ~\text{is} ~\hat{c},\\
			&\forall{x}\in \Gamma^+.
		\end{split}
	\end{equation}
	\begin{equation}\label{prior3}
		\begin{split}
			q^{-1}z\geq z_{sup} \Longrightarrow  &\text{The cost to misclassify sample }x ~\text{is} ~\hat{c},\\
			&\forall{x}\in \Gamma^+.
		\end{split}
	\end{equation}

	In other words, for the first classifier, $\mathcal{A}=\{x: q^{-1}z\leq z_{inf}\}$ and for the second classifier, $\mathcal{A}=\{x: q^{-1}z\geq z_{sup}\}$. We will substitute these piece of knowledge into the problem (\ref{modified svm3}) to model two real blast furnace datasets. Intuitively, prior Eq. (\ref{prior2}) can be understood as the fact that if the silicon content of the last record was (abnormally) low, then its current silicon content is possible to be still low. This is the same for prior Eq. (\ref{prior3}), which can be understood as the fact that if the silicon content of the last record was (abnormally) high, then its current silicon content is possible to be still high. This rule is not only coincident with the intuition of skilled operators of the blast furnace, but also coincident with the evolution of hot metal silicon content reflected in the real data as can be seen from section \Rmnum{5} . With the guidance of the prior knowledge, the black-box model can be expected to perform better.

	\subsection{Knowledge integrated into Soft-Margin SVM Model}
	As described previously, our goal is to build a classifier which can get some good leverage with the domain knowledge to make predictions that consistent with industrial requirements. Specifically, we hope to integrate the domain knowledge Eq. (\ref{prior2}) and Eq. (\ref{prior3}) with the black-box models optimally. In order to do this, our idea is to combine the prior knowledge just mentioned with the classifier to obtain a knowledge integrated model.
	
	Now, we try to build a knowledge incorporated model based on the idea of the model (\ref{svm2}).We also consider binary classification problem here. We set three parameters $c^{+}>0, c^{-}>0$ and $\hat{c}\geq 1$, which represents the cost for false negative, false positive and for false positive if the data also belong to the region $\mathcal{A}$, respectively. To be more specific, we assume the relationship between these three parameters in this context as, $c^{-}\geq c^{+}$. We hope that the derived decision hyperplane is more likely to identify the positive class data which lie in the region $\mathcal{A}$ correctly. We denote the complement of A as
	$\bar{\mathcal{A}}=\left\{x|g({x})> {0}\right\}$ here, so $\bar{\mathcal{A}}\cup \mathcal{A}=X$.

 Following the same idea of maximum-margin from the original SVM, if we denote $L(y_i,f(x_i)):=(1-y_{i}f(x_{i}))_{+}$, for each $i \in \left\{1,\dots ,m\right\}$, and we introduce a nonnegative variable $\xi_{ i }=\frac{c^{-}\hat{c}}{c^{+}}L(y_i,f(x_i))$, if $z_i \in R_1$, $\eta_{ i }=\frac{c^{-}}{c^{+}}L(y_i,f(x_i))$, if $z_i \in R_2$ and $\zeta_{ i }=L(y_i,f(x_i))$, if $z_i\in R_3$. Then, the model to assign different cost to different regions of the input space based on the prior knowledge mentioned in the previous subsection can be built to minimize the following objective function
	\begin{eqnarray}\label{modified svm2}
		\begin{array}{cl}
			\min\limits _{w,b,{\xi_i}\in R,{\eta_i}\in R,{\zeta_i}\in R} & \frac{1}{2C}w^Tw
			+\frac{1}{m}\left[\sum_{z_{i}\in R_1}\xi_{i}+\sum_{z_{j}\in R_2}\eta_{j}
			+\sum_{z_k \in R_3}\zeta_{k}\right]\\
			\text{s.t.} & y_{i}(w^T\Phi(\textbf{x}_i)+b)\geq 1-\frac{c^{+}}{c^{-}\hat{c}}\xi_{i},\quad z_{i}\in R_1\\
			& y_{i}(w^T\Phi(\textbf{x}_i)+b)\geq 1-\frac{c^{+}}{c^{-}}\eta_{i}, ~\quad z_{j}\in R_2\\
			& y_{i}(w^T\Phi(\textbf{x}_i)+b)\geq 1-\zeta_{i}, ~~~~~\quad z_k \in R_3\\
			& \xi_{i}\geq 0, ~\eta_j\geq 0, ~\zeta_{k}\geq 0, ~\quad \\
			&  i+j+k=m.
		\end{array} 
	\end{eqnarray}

 where $f\in{\mathcal{H}}_{K}$, $R_1:=\{z_i: x_{i} \in\mathcal{A} ~\&
	~y_i=+1\}$, $R_2:=\{z_i: x_{i} \in\bar{\mathcal{A}}~\&~y_i=+1\}$, $R_3:=\{z_i: y_i=-1\}$. Among them, $\textbf{z} = \left( z _ { i } \right) _ { i = 1} ^ { m }=\left( x _ { i } ,y _ { i } \right) _ { i = 1} ^ {m}=R_1\cup R_2\cup R_3$. Here, the regularization coefficients $\frac{c^{-}\hat{c}}{c^{+}}$, $\frac{c^{-}}{c^{+}}$, and 1 corresponding to regions $R_1$, $R_2$, and $R_3$, respectively, represent the misclassification cost of our model assigned to samples in different regions, reflecting the prior information mentioned earlier. According to the coefficient size relationship set at the beginning of this section, we see that the model believes that the sample in region $R_1$ has the highest misclassification cost, the sample in region $R_2$ is second, and the sample in region $R_3$ ranks last, which is consistent with the prior knowledge. As will be seen later, this fits our application to the blast furnace context. This optimization problem can be further re-written as a constrained optimization problem with a differentiable objective function. 
 
	We can see that different types of samples correspond to different constraints (or different soft-margins) in this model (\ref{modified svm2}) (hereinafter referred to as the \textbf{knowledge integrated SVM}).

    The proposed model (\ref{modified svm2}) doesn't need to sample extra examples outside the original training set when solving it. Therefore, our knowledge integrated SVM has the same computation load as the standard SVM.

	To solve the proposed knowledge integrated model (\ref{modified svm2}), we first transform it into the Lagrange dual problem as follow. Here, both classifiers correspond to the following model
	\begin{eqnarray}\label{modified svm3}
		\begin{array}{cl}
			\min_{\alpha} & \frac{1}{2}\sum_{i=1}^{m}\sum_{j=1}^{m}\alpha_i\alpha_jy_iy_jK(x_i,x_j)-\sum_{i=1}^{m}\alpha_i\\
			
			\text{s.t.} & \sum_{i=1}^{m}\alpha_iy_i=0 \\
			& 0\leq \alpha_i \leq {c^{-}\hat{c}}, ~~~~ x_{i}\in\mathcal{A} ~\&~ y_{i}=1 \\
			& 0\leq \alpha_i \leq {c^{-}}, ~~~~~x_{i}\in\bar{\mathcal{A}} ~\&~ y_{i}=1\\
			& 0\leq \alpha_i \leq {c^{+}}, \quad~~  y_{i}=1.
		\end{array} 
	\end{eqnarray}
 Here, we omit the common multiplier $\frac{C}{m}$ of the upper bounds on $\alpha$ in the second, third, and the last constraints. $K(\cdot, \cdot)$ is defined to be the Gaussian radial basis kernel with the form: $K(x_i,x_j)=\exp\left({-\gamma\|x_i-x_j\|^2}\right)$. 

    Finally, we can solve this problem (\ref{modified svm3}) using the MATLAB code where the gird search together with five-fold cross validation techniques are used to train the model parameters $ \{\gamma, c^{-},c^{+},\hat{c}\} $. Also, we assume the size order among the penalty parameters $\{c^-,c^+,\hat{c}\}$ as $c^{-}\geq c^{+}$ and $\hat{c}\geq 1$, based on which we divide the data into three groups (samples with negative labels, samples with positive labels lie in the region $\mathcal{A}$ and the rest of positive samples). We set the grid search ranges of these parameters, include $\gamma$, $\hat{c}$, $c^{-}$, $c^{+}$. The quantity value of each penalty parameter indicates the importance of each class of samples to be classified correctly.
    
	It should be pointed out that from a theoretical point of view, the model of (\ref{modified svm2}) does not contribute much compared to the work of Lin et al.\cite{lin2002support} and Wu et al.\cite{wu2004incorporating}. However, the main contribution of this paper is the acquisition of domain knowledge in (\ref{prior1}), which is usually readily available in the context of a specific problem. Furthermore, incorporating domain knowledge into the model helps to increase the transparency of the black-box SVM model, as through domain knowledge integrated into model it requires the model to classify as accurately as possible samples in regions of particular interest to us. Of course, the knowledge integrated SVM model in (\ref{modified svm2}) cannot be considered as a white-box model either, it is just a partially transparent black-box model, which artificially affects the classification results of the classifier, but the output results of the model still cannot be accurately predicted.  We leave this point as a future study.
	
	\subsection{Measurement of the Blast Furnace Classification Model}
	Here, we measure the "performance" of a classification model in the sense of the \textbf{ensemble accuracy}, which is induced from the practical industry goals. We define the ensemble accuracy of a classification model in the context as the weighted sum of test accuracies for different classes of samples. As mentioned before, the importance of each class of samples
	that need to be classified right are assigned based on the corresponding penalty parameters $\{c^-,c^+,\hat{c}\}$. Specifically, we quantify the ensemble accuracy of any classifier $f$ to be
	\begin{eqnarray}
		\text{Ensemble}\text{Acc}(f)&:=&0.6\cdot\text{Acc}_\text{class1}+0.1\cdot\text{Acc}_\text{class2}\nonumber\\
		&+&0.3\cdot\text{Acc}_\text{class3},
		\label{ensemble}
	\end{eqnarray}
	where we separate samples into three classes as, $\text{class1}:=\{z_i: x_i\in\mathcal{A} ~\& ~y_i=+1\}$, $\text{class2}:=\{z_j: x_j\in\bar{\mathcal{A}} ~\& ~y_j=+1\}$, and $\text{class3}:=\{z_k: y_k=-1\}$. The test accuracy of $\text{class1}$ occupies the most significant ratio of the ensemble accuracy, which is due to that the misclassification of samples from $\text{class1}$ is more dangerous than that of the other two classes of examples (more likely results in the happening of the chilled hearth and high energy consumption). In other words, the presented ensemble accuracy can be seen as a criterion that not only focuses on the accuracy of identifying abnormally low or high silicon samples, but also the accuracy of classifying non-low or proper silicon samples for the first or second classifier.

	\section{CASE STUDY: TWO REAL BLAST FURNACES}
	In this section, the knowledge integrated SVM model in (\ref{modified svm2}) is used to model 2 real blast furnace datasets, and solve a three-class classification problem as designed in Section \Rmnum{2}. The domain knowledge for each dataset is derived from the silicon evolution in real blast furnaces. Our numerical results reveal that it is possible to improve the performance of SVM models in \cite{lin2002support} by incorporating domain knowledge into them.

	\subsection{Blast Furnace Data Collection} 
	
	Our experimental data are collected from two typical Chinese blast furnaces with the inner volume of about $2500 ~\text{m}^3$ and $750 ~\text{m}^3$, which are labeled as the blast furnaces (a) and (b), respectively. Tables I and II present the selected 7 input variables for these two blast furnaces. Notice that some lagged terms are also treated as the inputs which are due to the ($2-8\text{h}$) time delay for the blast furnace outputs responding to the inputs. Cause the sampling interval is about $1.5\text{h}$ for the blast furnace (a) while $2\text{h}$ for the blast furnace (b), we determine the time delay for the variables of these two blast furnaces as $5$ and $4$, respectively.
	
\begin{figure}[!t]\centering
		\includegraphics[width=8.5cm]{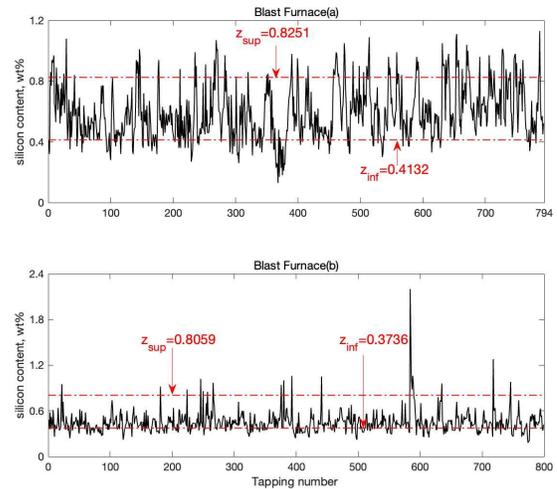}
		\caption{Silicon content evolution of the blast furnaces (a) and (b)}\label{silicon_change}
	\end{figure}

	We sample $794$ and $800$ records for the blast furnaces (a) and (b), respectively. Among them, for each random experiment, we choose $500$ points as the training set and $100$ records from the remained dataset as the testing set for the blast furnace (a) and (b). We performed 100 random experiments to confirm the effectiveness of our model. Fig. \ref{silicon_change} shows the silicon evolution of the two blast furnaces. According to the range of silicon classification obtained by clustering five centers in \cite{gao2014rule}, we set the acceptable infimum and supremum of $z$ as $z_{inf}^{a}=0.4132/z_{sup}^{a}=0.8251$ for the blast furnace (a), and $z_{inf}^{b}=0.3736/z_{sup}^{b}=0.8059$ for the blast furnace (b) by referring to the results from \cite{gao2014rule}. Then, in all sample points, we get $115/216$ ``low silicon" samples for the blast furnace (a)/(b),  $569/564$ ``proper silicon" samples for the blast furnace (a)/(b) and $110/20$ ``high silicon" samples for the blast furnace (a)/(b).

		\begin{table}[!t]
		\renewcommand{\arraystretch}{1.3}
		\setlength{\abovecaptionskip}{0pt}
		\setlength{\belowcaptionskip}{0pt}
		\caption{\textsc{Relevent input variables of Blast Furnace} (a)}
		\label{Tab_1} 
		\centering
		\resizebox{\columnwidth}{!}{
			\begin{tabular}{l l l}
				
				\toprule
				\multicolumn{1}{c}{Variable name [Unit]} & \multicolumn{1}{c}{Symbol} & \multicolumn{1}{c}{\pbox{20cm}{Input variable }}  \\
				\midrule
				Blast temperature  [$^\circ \mathrm{C}$] & $x^{(1)}$ & ${q^0}^\S$,$q^{-1}$,$q^{-2}$,$q^{-3}$,$q^{-4}$,$q^{-5}$\\
				Blast volume       [m$^3$/min]   & $x^{(2)}$ & $q^0$,$q^{-1}$,$q^{-2}$,$q^{-3}$,$q^{-4}$,$q^{-5}$\\
				Feed speed         [mm/h]   & $x^{(3)}$ & {$q^0$,$q^{-1}$},$q^{-2}$,$q^{-3}$,$q^{-4}$,$q^{-5}$\\
				Gas permeability   [m$^3$/min$\cdot$kPa]  & $x^{(4)}$ & $q^0$,$q^{-1}$,{$q^{-2}$},$q^{-3}$,$q^{-4}$,$q^{-5}$\\
				Pulverized coal injection  [ton]  & $x^{(5)}$ & {$q^0$,$q^{-1}$},{$q^{-2}$,$q^{-3}$},{$q^{-4}$},$q^{-5}$\\
				Sulfur content     [wt\%]   & $x^{(6)}$ & $q^{-1}$\\
				Silicon content    [wt\%]  & $z$ & {$q^{-1}$}\\ 
				
				\bottomrule\multicolumn{3}{l}{\hspace{-2mm}\scriptsize{$^\S$
						$q^0,q^{-1},\cdots,q^{-5}$ represent delay operators, such as
						$q^{-1}x(t)=x(t-1)$. }}
		\end{tabular}}
	\end{table}
	
	\begin{table}[!t]
		\renewcommand{\arraystretch}{1.3}
		\setlength{\abovecaptionskip}{0pt}
		\setlength{\belowcaptionskip}{0pt}
		\caption{\textsc{Relevent input variables of Blast Furnace} (b)} \label{Tab_2} 
		\centering
		\resizebox{\columnwidth}{!}{
			\begin{tabular}{l l l}
				
				\toprule
				\multicolumn{1}{c}{Variable name [Unit]} & \multicolumn{1}{c}{Symbol} & \multicolumn{1}{c}{\pbox{20cm}{Input variable }}  \\ \hline
				Blast temperature  [$^\circ \mathrm{C}$] & $x^{(1)}$ & ${q^0}$,$q^{-1}$,$q^{-2}$,$q^{-3}$,$q^{-4}$\\
				Blast volume       [m$^3$/min]   & $x^{(2)}$ & $q^0$,$q^{-1}$,$q^{-2}$,$q^{-3}$,$q^{-4}$\\
				Feed speed         [mm/h]   & $x^{(3)}$ & {$q^0$,$q^{-1}$},$q^{-2}$,$q^{-3}$,$q^{-4}$\\
				Gas permeability   [m$^3$/min$\cdot$kPa]  & $x^{(4)}$ & $q^0$,$q^{-1}$,{$q^{-2}$},$q^{-3}$,$q^{-4}$\\
				Pulverized coal injection  [ton]  & $x^{(5)}$ & {$q^0$,$q^{-1}$},{$q^{-2}$,$q^{-3}$},{$q^{-4}$}\\
				Sulfur content     [wt\%]   & $x^{(6)}$ & $q^{-1}$\\
				Silicon content    [wt\%]  & $z$ & {$q^{-1}$}\\
			
				\bottomrule\multicolumn{3}{l}{\hspace{-2mm}}
		\end{tabular}}
	\end{table}
 
\subsection{Reliability of Blast Furnace Prior Information }
In the section \Rmnum{4}, we introduce the prior information and knowledge integrated SVM for classification problems, the accuracy of the prior information about Blast Furnace can be verified by the real data of blast furnace (a) and (b).

\begin{figure}[!t]\centering
		\includegraphics[width=8cm]{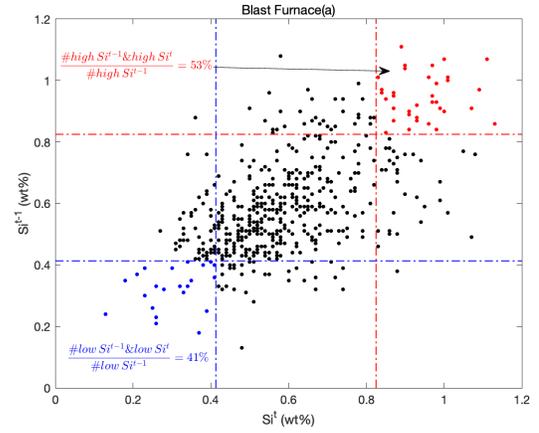}
		\caption{Reliability of blast furnace (a) prior information}\label{a_motivation}
	\end{figure}

\begin{figure}[!t]\centering
		\includegraphics[width=8cm]{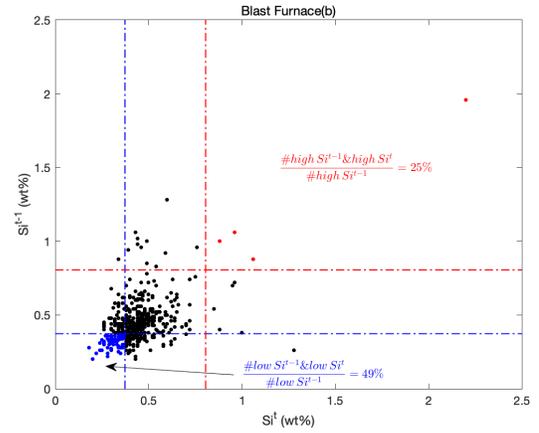}
		\caption{Reliability of blast furnace (b) prior information}\label{b_motivation}
	\end{figure}

Figure \ref{a_motivation} and figure \ref{b_motivation} is scatter diagrams of random training datasets of blast furnace (a) and (b) (both include 500 data points). The abscissa represents the silicon content information at this moment, and the ordinate represents the silicon content information at the previous moment. These figures reflect the reliability of the prior information of the two blast furnaces. Among them, $\#$ indicates the card of the set, i.e., the number of elements in the set, and low ${Si}^{t-1}$, low ${Si}^{t}$, high ${Si}^{t-1}$, high ${Si}^{t}$ respectively indicates the low silicon at the previous moment, the low silicon at this moment, the high silicon at the previous moment and the high silicon at this moment. The blue solid in the figure indicates the sample points that were low silicon at the previous moment and are also low silicon at this moment. The formula with the color blue is the proportion of these sample points to the low silicon samples at the previous moment. The red solid and the red formula also reflect the same information for the high silicon samples. We can see that for blast furnace (a)/(b), $$ \frac{\#\left\{{\rm low} \,{Si}^{t-1} \& \,{\rm low} \,{Si}^{t}\right\}}{\#\left\{{\rm low} \,{Si}^{t-1}\right\}} $$ is 41\%/49\%, and $$ \frac{\#\left\{{\rm high} \,{Si}^{t-1} \& \,{\rm high} \,{Si}^{t}\right\}}{\#\left\{{\rm high} \,{Si}^{t-1}\right\}} $$ is 53\%/25\%. Except for the high silicon content of blast furnace (b), the values are all close to 50\%, which reflects the statistical law of the evolution of silicon content in blast furnace hot metal and also reflects the plausibility of the prior information we adopt in Section \Rmnum{4}. We will see from the experimental results later that the high silicon information of blast furnace (b) can also improve the classification accuracy of high silicon samples, so the high silicon related prior information of blast furnace (b) is also useful for improving the classification model.
 
	\subsection{Experimental Results Discussion}
	As can be seen from Section \Rmnum{2}, the blast furnace problem involves three classes of silicon classification tasks, which we perform here by a one-to-many approach \cite{vapnik1998statistical}. For each dataset, we design two binary classifiers. For blast furnace (a) and (b), the first classifier is used to distinguish "low silicon" samples from the other two types and we treat low silicon as the positive class, while the second classifier is used to distinguish "proper silicon" and "high silicon" samples and we treat high silicon as the positive class. 
\begin{figure*}[!t]
		\centering
		\includegraphics[width=19cm]{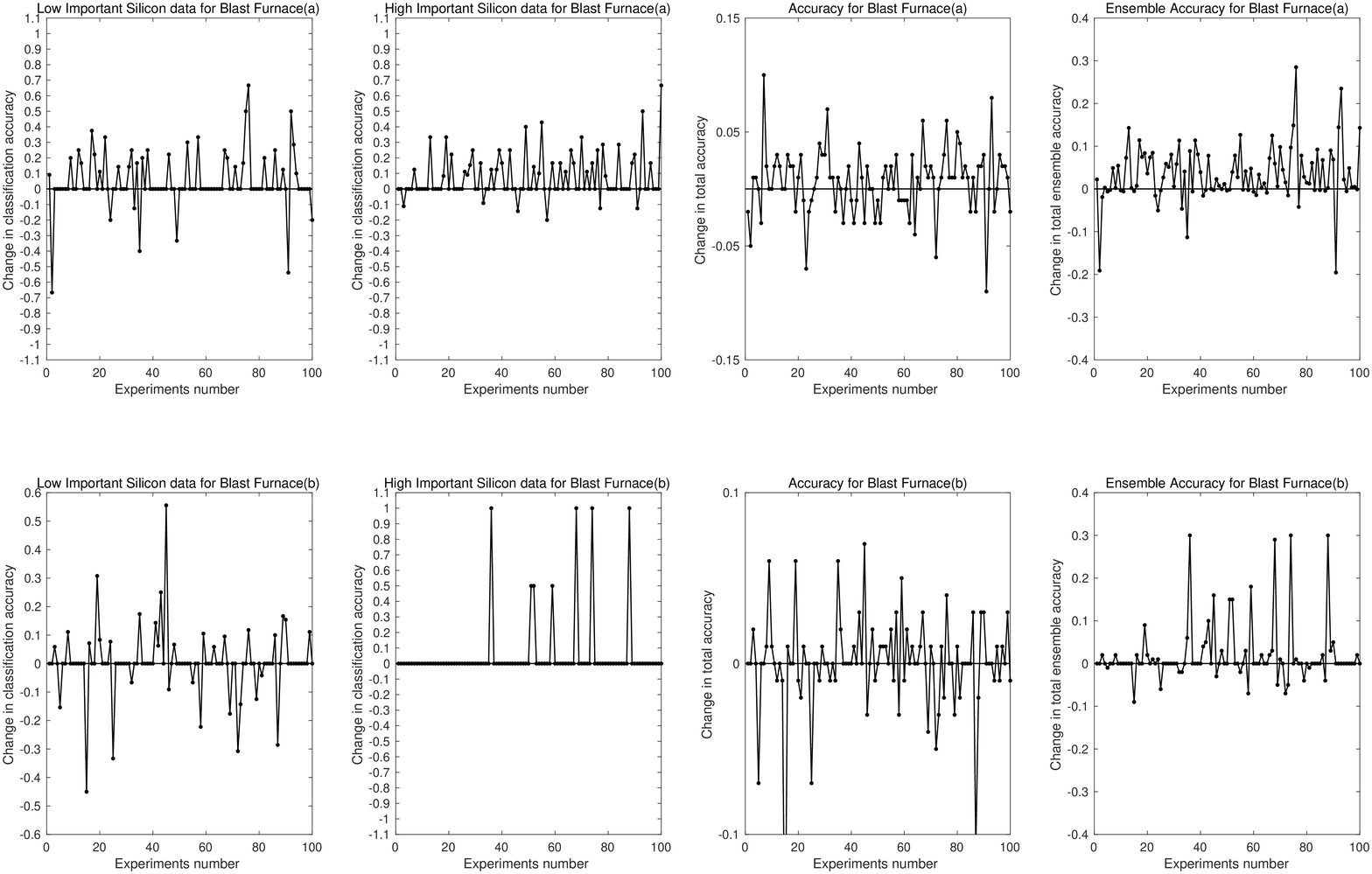}
		\caption{The result of the difference value in classification accuracy of low important samples,  high important samples,  accuracy and ensemble accuracy between model (\ref{modified svm3}) and model (\ref{svm2}) in 100 random experiments for the blast furnace (a) and (b).}
		
		\label{eight}
	\end{figure*}
	
 In all experiments, we randomly select 600 samples from the dataset, and select the first 500 samples as the training dataset and the remaining 100 as the testing set, and this process is repeated 100 times to observe the reliability of the results. The training set is used to learn the model parameters, while the testing set is used to evaluate the performance of our model (\ref{modified svm3}). We tune the hyperparameters through a grid search together with  five fold cross-validation. All experiments were performed in Matlab9.12 runs on an environment with a 3.2GHz M1 pro processor and 32.00 GB memory. The grid search intervals for the models are set as following. For the model (\ref{modified svm2}), $\{\gamma,\hat{c}, c^{-}, c^{+}\}$   are set to be $\gamma = \{2^{-4}, 2^{-2}, 2^0, 2^2, 2^4\}$, $\hat{c}= \{1, 2, 3, 4, 5\}$, $c^{-}= 2$ and $c^{+}= 1$. For the model (\ref{svm2}), $\{\gamma, c^{-}, c^{+}\}$ are set to be $\gamma = \{2^{-4}, 2^{-2}, 2^0, 2^2, 2^4\}$, $c^{-} = \{1, 2, 3, 4, 5\}$ and $c^{+}= 1$. For the blast (a) and (b), and for both classifiers, the parameter grid search interval is the same. 
 \begin{table*}[!t] 
		\renewcommand{\arraystretch}{1.6}
		\setlength{\tabcolsep}{5pt}
		\setlength{\abovecaptionskip}{0pt}
		\setlength{\belowcaptionskip}{0pt}
		\caption{\textsc{Experimental Results of Blast Furnace} (a) \textsc{and} (b)}
		\label{Tab_3} 
		\centering
			\scalebox{0.8}{
				\begin{tabular}{ c c c c c c c c c }
				 \\[-3mm]
				\toprule
					\multicolumn{1}{c}{ Blast Furnace } &
					\multicolumn{1}{c}{ Average Accuracy(\%)} & \multicolumn{1}{c}{low} & \multicolumn{1}{c}{\pbox{20cm}{low important }} & \multicolumn{1}{c}{proper}& \multicolumn{1}{c}{high}& \multicolumn{1}{c}{high important}& \multicolumn{1}{c}{Accuracy}& \multicolumn{1}{c}{Ensemble accuracy} \\[1.6ex] 
				\midrule
				   \multirow{2}{*}{(a)}&	Model with Knowledge Integrated  & \textbf{43.53} &	\textbf{68.77}	& \textbf{93.97} &	\textbf{44.92}	& \textbf{78.56} &	\textbf{79.72} &	\textbf{72.74} \\
				~ &	Model without Knowledge Integrated & 42.74 &	64.09 &	93.30 	& 44.03 &	71.40 &	79.13 	& 69.36 \\ 
						\midrule
					 \multirow{2}{*}{(b)}&
						Model with Knowledge Integrated  & \textbf{37.10} &	\textbf{52.37} &	97.63 &	\textbf{30.89} &	\textbf{51.50} &	79.77 	& \textbf{59.19}  \\
					~&	Model without Knowledge Integrated & 37.08 	& 51.96 &	\textbf{97.70} 	& 29.35 &	46.00 &	\textbf{79.81} &	57.22  \\ 
						
						\bottomrule\multicolumn{3}{l}{\hspace{-2mm}}
				\end{tabular}}
			\end{table*}	
 
  Figure \ref{eight} shows the change in classification accuracy in 100 random experiments for blast furnace (a) and blast furnace (b), that is, accuracy of the knowledge integrated SVM minus the accuracy of the without knowledge integrated SVM. Here, we focus on four accuracy rates, which are low importance sample classification accuracy, high importance sample classification accuracy, accuracy, and ensemble accuracy between model (\ref{modified svm2}) and model (\ref{svm2}). Among them,  "accuracy" is common classification accuracy measure, that is, the number of correctly classified samples in the testing set is divided by the total number of samples, and "ensemble accuracy" is calculated using  Eq.(\ref{ensemble}) for the two classifiers, and then averaged to obtain the result which is used as the final ensemble accuracy of the blast furnace testing set. The small solid black circle in the figure represent the results of each test. 
 
 In Figure \ref{eight}, since the vertical axis reflects the change in classification accuracy between model (\ref{modified svm2}) and model (\ref{svm2}), the more solid black dots on the upper half plane, the more experiments with higher accuracy of our model in 100 experiments. The farther a solid point on the upper half plane is from the horizontal axis, the more the accuracy of our model improves compared to model (\ref{svm2}) in the corresponding experiment. 
 First of all, we can see that in one hundred random experiments, our model is more accurate than the model without knowledge integrated in classifying high importance silicon samples in most experiments, and the improvement rate is basically above 0.1. But our model seems to be better than svm without knowledge integrated only on blast furnace (a). This result may be related to the division of training and testing sets. From table \ref{Tab_3}, we can find that our model still performs better in classification of low important samples of blast furnace (b).
 Secondly, through the observation of the accuracy and  ensemble accuracy, we can see that for blast furnace (a) and blast furnace (b), the accuracy of the two models is comparable, indicating that our model does not reduce the overall prediction accuracy, while the ensemble accuracy of our model is superior, which is the accuracy measure we mainly focus on in actual industry. This result shows that our model is more helpful to actual blast furnace problem. Therefore, the test results of 100 times in the figure intuitively reflect the effectiveness of our model in classifying important samples, as well as the help to actual industrial targets. In order to quantitatively analyze the results of the 100 random experiments as a whole, we will then compare and analyze the experimental results in the average sense.

	Table \ref{Tab_3} exhibits the performance of our knowledge integrated model (\ref{modified svm2}) comparing to that of the baseline model (\ref{svm2}) in the average sense. We can see that the concerned accuracies of the model (\ref{svm2}) for two blast furnace datasets are both improved after incorporating prior knowledge into it. The ensemble accuracy increases from $69.36\%$ to $72.74\%$ for the blast furnace (a), and $57.22\%$ to $59.19\%$ for the blast furnace (b). Moreover, higher test accuracies with respect to the low/high important samples are obtained, $68.77\%/ 52.37\%$ v.s. $64.09\%/ 51.96\%$ and $78.56\%/ 51.50\%$ v.s. $71.40\%/ 46.00\%$ for the blast furnace (a) and (b), respectively. This result indicates that the proposed knowledge integrated method is more effective than the original one in identifying the focused class of data (some important "low/high silicon" data that should be identified correctly).

	We also compare the accuracies on the "low silicon" and "high silicon" datasets yielded by these two models in table \ref{Tab_3}. We can observe that the average accuracy of the presented knowledge integrated model (\ref{modified svm2}) on the "low silicon" samples is better than that of the baseline model (\ref{svm2}), about $0.79\%$/$0.02\%$ increasement for the blast furnace (a)/(b), respectively. Also the average accuracy of the presented knowledge integrated model (\ref{modified svm2}) on the "high silicon" samples is better than that of the baseline model (\ref{svm2}), about $0.89\%$/$1.54\%$ increasement for the blast furnace (a)/(b), respectively. Traditional classification models usually tend to have the higher precision on the class of samples with a large proportion and the lower precision on the class of samples with fewer size, when handling the unbalanced data. However, our results (see Fig. \ref{eight}) indicate that our knowledge integrated learning scheme (\ref{modified svm2}) not only performs better than that of model (\ref{svm2}) in identifying the class of data with a low proportion (the "low/high silicon"  data), but also maintains the high accuracy on separating the class of data with a large proportion (the "proper silicon" data). Of course, as can be seen from the average of the accuracy, the overall prediction performance of our model are also comparable.
			
	To make further rigorous comparison, we also make a paired student's t-test on the ensemble accuracy results yielded by these two models, and get the p-values to be less than 0.05 for the blast furnace (a) and (b). So when they are applied to model these two blast furnaces, it is of high possibility for the model (\ref{modified svm2}) to be more effective than the baseline model (\ref{svm2}).
			
	In summary, incorporating prior knowledge into the black-box classification model is an efficient way to guide the model to perform better based on some specific performance measure, i.e., the ensemble accuracy. Thus, our knowledge incorporated learning framework provides a possibility to interact human knowledge with the black-box models for achieving the desired performance in real applications, i.e., the safety and energy saving requirement on the blast furnace problems. Intuitively, the integrated prior knowledge acts as an auxiliary constraint for imposing some regularities on the black-box models. In our example, (\ref{prior2}) owns the potential of teaching the black-box model to learn the silicon evolution pattern (that is, if the silicon content of the last record was low or high, then its current silicon content is possible to keep low or high). Therefore, it makes the knowledge integrated model perform better in the sense of the ensemble accuracy.
			
\section{Conclusion}
   In this paper, we propose a method to design the classifier that satisfys the desired learning target. We provide a meaningful strategy for modeling with the knowledge incorporated. Further, the experimental results on two real blast furnace datasets indicate that our knowledge incorporated learning scheme performs better than the original model when applied to the two real blast furnace dataset. Therefore, our method offers the domain experts an useful way to interact with the black-box models to achieve better performance.
			
	The main trick in this paper is to integrate prior knowledge with the learning target from which the knowledge based SVM classifier is derived. Therefore, our knowledge incorporated method is new.
			
	The present work can be generalized to other blast furnace datasets to continue using this model by combining domain knowledge with classifiers for different learning targets. In addition, from experiments we see that our model is also suitable for imbalanced datasets.
 Finally, other black-box models can also incorporate domain knowledge into the model through similar methods to obtain the desired performance.
			
\section*{Acknowledgment}
			
	We would like to acknowledge support for this project from the National Natural Science Foundation of China under Grant No. 11671418 and 61611130124.

\bibliographystyle{IEEEtranTIE}
			
\bibliography{IEEEabrv,ref}

 \end{document}